\documentclass{article}

\usepackage{times}
\usepackage{graphicx} 
\usepackage{subfigure} 
\usepackage{microtype}
\usepackage{todonotes}
\usepackage{bbm}
\usepackage{amsfonts}
\usepackage{amsmath}
\usepackage{natbib}

\usepackage{algorithm}
\usepackage{algorithmic}

\usepackage{hyperref}

\newcommand{\cut}[1]{}

\usepackage[accepted]{whi2016}

\DeclareMathOperator*{\argmin}{argmin}

\def\Sim{\Pi}
\def\EModel{\xi}

\icmltitlerunning{Model-Agnostic Interpretability of Machine Learning}

\begin{document} 

\twocolumn[
\icmltitle{Model-Agnostic Interpretability of Machine Learning}

\icmlauthor{Marco Tulio Ribeiro}{marcotcr@cs.uw.edu}
\icmlauthor{Sameer Singh}{sameer@cs.uw.edu}
\icmlauthor{Carlos Guestrin}{guestrin@cs.uw.edu}
\icmladdress{University of Washington
            Seattle, WA 98195 USA}

\icmlkeywords{interpretability, machine learning, comprehensibility, interactive}

\vskip 0.3in
]

\begin{abstract} 
Understanding why machine learning models behave the way they do empowers both system designers and end-users in many ways: in model selection, feature engineering, in order to \emph{trust} and act upon the predictions, and in more intuitive user interfaces.
Thus, interpretability has become a vital concern in machine learning, and work in the area of interpretable models has found renewed interest.
In some applications, such models are as accurate as non-interpretable ones, and thus are preferred for their transparency. 
Even when they are not accurate, they may still be preferred when interpretability is of paramount importance.
However, restricting machine learning to interpretable models is often a severe limitation.
In this paper we argue for explaining machine learning predictions using \emph{model-agnostic} approaches. 
By treating the machine learning models as black-box functions, these approaches provide crucial flexibility in the choice of models, explanations, and representations, improving debugging, comparison, and interfaces for a variety of users and models.
We also outline the main challenges for such methods, and review a recently-introduced model-agnostic explanation approach (LIME) that addresses these challenges.
\end{abstract} 

\section{Introduction}

As machine learning becomes a crucial component of an ever-growing number of user-facing applications, \emph{interpretable machine learning} has become an increasingly important area of research for a number of reasons.
First, as humans are the ones who train, deploy, and often use the predictions of machine learning models in the real world, it is of utmost importance for them to be able to trust the model.
Apart from indicators such as accuracy on sample instances, a user's trust is directly impacted by how much they can understand and predict the model's behavior, as opposed to treating it as a black box.
Second, a system designer who understands why their model is making predictions is certainly better equipped to improve it by means of feature engineering, parameter tuning, or even by replacing the model with a different one.
Lastly, even in lower stakes domains such as movie or book recommendations, getting a rationale such as ``you will probably like this book because of your interest in Russian Literature'' makes the model much more useful to the users, and more likely to be trusted.
Thus there is a crucial need to be able to explain machine learning predictions, i.e. provide users a rationale for why a prediction was made using textual and visual components of the data, and/or producing counter-factual knowledge of what would happen were the components different.

\cut{
Machine learning is at the core of many recent advances in science and technology.
Unfortunately, the important role of humans is an oft-overlooked aspect in the field, with many of the state of the art models being functionally black boxes.

Humans are always the ones who train and deploy machine learning models, and very often are the final consumers of its predictions.
Among other indicators such as accuracy and example predictions, a user's trust is directly impacted by how much they understand the model's behavior, as opposed to seeing it as a black box.

Among other indicators such as accuracy and example predictions, a user's trust is directly impacted by how much they understand the model's behavior, as opposed to seeing it as a black box.
(2) In a similar vein, an analyst that understands why his model is making certain predictions is certainly better empowered to improve such a model by means of feature engineering, parameter tuning, or even by replacing it by a more (or less) flexible model.
(3) Finally, even in lower stakes domains such as movie or book recommendations, getting rationales such as ``you will probably like this book because of your interest in Russian Literature'' makes the model much more useful to the users, and more likely to be trusted.

We now define the problem of ``explaining a prediction'' loosely as presenting textual or visual artifacts that provide qualitative understanding of the relationship between an instance's components (e.g. words in text, patches in an image) and the model's prediction.
An explanation can give the human a rationale for why a prediction was made (e.g. in terms of contributions from the components), and/or the counter-factual knowledge of what would happen were the components different.
In a similar fashion, ``explaining a model'' is presenting artifacts that provide qualitative understanding of how the model behaves in general.
One way of explaining a model is by explaining a subset of its predictions.
}

The prevailing solution to this explanation problem is to use so called ``interpretable'' models, such as decision trees, rules~\cite{LethamRuMcMa15, WangRu15}, additive models~\cite{caruana2015}, attention-based networks~\cite{Xu2015show}, or sparse linear models~\cite{supersparse}.
Instead of supporting models that are functionally black-boxes, such as an arbitrary neural network or random forests with thousands of trees, these approaches use models in which there is the possibility of meaningfully inspecting model components directly --- e.g. a path in a decision tree, a single rule, or the weight of a specific feature in a linear model. 
As long as the model is accurate for the task, and uses a reasonably restricted number of internal components (i.e. paths, rules, or features), such approaches provide extremely useful insights.


An alternative approach to interpretability in machine learning is to be \emph{model-agnostic}, i.e. to extract post-hoc explanations by treating the original model as a black box.
This involves learning an interpretable model on the predictions of the black box model~\cite{craven,Baehrens:2010:EIC:1756006.1859912}, perturbing inputs and seeing how the black box model reacts~\cite{gametheory, krauseinteracting}, or both~\cite{lime}. 

In this position paper, we argue for separating explanations from the model (i.e. being model agnostic).
The summary of our position is that restricting the space of models to be interpretable is a constraint that results in less flexibility, accuracy, and usability.
We develop this position with examples, while also describing the inherent challenges in model agnosticism. 
Finally, we review the recently-introduced LIME approach~\cite{lime}, and discuss how it provides many of the desirable characteristics for model-agnostic explanations.


\section{A Case for Model Agnosticism}
In this section, we make a case for model-agnostic interpretability, as opposed to just using interpretable models.


\subsection{Model Flexibility}


For most real-world applications, it is necessary to train models that are accurate for the task, irrespective of how complex or uninterpretable the underlying mechanism may be. 
We can observe this ideology manifesting with the increasing commonplace deployment of uninterpretable deep neural architectures for a wide variety of tasks.

Interpretable models for such tasks remain unsatisfying; such models are inherently crippled by the need to be understandable, being susceptible to the limited ``perception budget''~\cite{magicnumber7} of the users.
This trade-off between model flexibility and interpretability \cite{freitas} implies one cannot use a model whose behavior is very complex, yet expect humans to fully comprehend it globally.
For example, for a task such as predicting the sentiment of a sentence, producing an accurate model that is understandable seems like an unfeasible task.
The size of the vocabulary alone makes it impossible for a short set of rules, a decision tree, or an additive model to be sufficiently accurate, not to mention more complex word interactions such as negation. 
Tasks that involve sensory data, such as audio and images, also suffer from the same problem: for a model to be useful, it must be sufficiently flexible to handle the data complexity.


In model-agnostic interpretability, the model is treated as a black box.
The separation of interpretability from the model thus frees up the model to be as flexible as necessary for the task, enabling the use of any machine learning approach - including, for example, arbitrary deep neural networks.
It also allows for the control of the complexity-interpretability trade-off (see next section), or ``failing gracefully'' if an interpretable explanation is not possible.

\cut{
 and thus can be as flexible as needed.
Model-agnostic approaches usually focus on explaining individual predictions, which is a more feasible task than explaining the model globally, even for very complex models.
It is easy to explain why sentences such as ``This is not bad.'' have a positive sentiment, even if we are not able to explain the whole sentiment prediction model.
Likewise, explaining why a particular picture contains a cat is simpler than explaining a cat-detection model globally.
Finally, by being model-agnostic one can control the complexity-interpretability trade-off (see next section), and ``fail gracefully'' if an interpretable explanation is not possible.
}

\subsection{Explanation Flexibility}
Different kinds of explanations meet different information needs.
In some cases, users may only care about positive evidence towards a certain prediction (e.g. which part of an image is most responsible for the prediction), while in other instances knowing the negative evidence may be useful (e.g.  in debugging a classifier).
Yet in other cases, the information need may be of counter-factuals, e.g. how the model would behave if certain features had different values.
Different users may also be able to handle different kinds of explanations; a user trained in statistics may be able to understand a Bayesian network, while a linear model is more intuitive to the layman.
Even if the explanation type is kept fixed, users may tolerate different granularities in different situations.
For example, \citet{freitas} notes a case where 41 rules are considered overwhelming, and contrasts it to another user who patiently analyzed 29,050 rules.

Most interpretable models are, however, restricted in what explanations are possible, be it a prototype~\cite{bayesian_case_model}, a set of rules~\cite{LethamRuMcMa15} or line graphs~\cite{caruana2015}.
Further, other constraints on interpretability, such as granularity, also have to be set \emph{a~priori} (e.g. max number of rules).
On the other hand, by keeping the model separate from the explanations, one is able to tailor the explanation to the information need, while keeping the model fixed.
If it is possible to measure how faithful the explanation is to the original model, one can effectively control the trade-off between fidelity and interpretability, as favored by \citet{freitas}.
Such approaches may also be able to provide multiple explanations of different types to the user, perhaps automatically picking the one with the highest faithfulness. 
Thus, by being model-agnostic, the same model can be explained with different types of explanations, and different degrees of interpretability for each type of explanation. 


\subsection{Representation Flexibility}
In domains such as images, audio and text, many of the features used to represent instances in state-of-the-art solutions are themselves not interpretable.
Unsupervised feature learning produces representations such as word embeddings~\cite{wordvec}, or the so-called deep features~\cite{deepfeatures}.
While an interpretable model trained on such features is still uninterpretable, model-agnostic approaches can generate explanations using different features than the one used by the underlying model.
Thus, even if the model is using word embeddings, the explanations can be in terms of words, for example.

\subsection{Lower Cost to Switch}
Switching models is not an uncommon operation in machine learning pipelines.
If one commits to using an interpretable model, one is ``locked-in'' to a particular model and a particular kind of explanations - even if newer, more accurate models are developed.
Even when the switch is from one interpretable model to another, users may have to be re-trained in understanding the new explanations, and the model's utility may decrease due to cognitive overhead.
In contrast, if one uses model-agnostic explanations, switching the underlying model for a new one is trivial, while the way in which the explanations are presented is maintained.

\subsection{Comparing Two Models}
When deploying machine learning in the real world, a system designer often has to decide between one or more contenders, and an incumbent model.
This comparison is hard to do if any of the systems are using interpretable models, while others are not.
Further, even if all of the models are interpretable, it may still be difficult to compare the insights gained from each if the underlying explanations are different in their representation - for example comparing a rule-based model with a tree-based model.
It is also not clear what to do if one of the contenders is less accurate but more interpretable, or vice versa.
With model-agnostic explanations, the models being compared can be explained using the same techniques and representations.

\section{Challenges for Model-agnostic Explanations}
While we have made a case for model agnosticism, this approach is not without its challenges.
For example, getting a global understanding of the model may be hard if the model is very complex, due to the trade-off between flexibility and interpretability. To make matters worse, local explanations may be inconsistent with one another, since a flexible model may use a certain feature in different ways depending on the other features.
In \citet{lime} we explained text models by selecting a small number of representative and non-redundant individual prediction explanations obtained via submodular optimization, similar in spirit to showing prototypes~\cite{bayesian_case_model}.
However, it is unclear on how to extend this approach to domains such as images or tabular data, where the data itself is not sparse.

In some domains, exact explanations may be required (e.g. for legal or ethical reasons), and using a black-box may be unacceptable (or even illegal).
Interpretable models may also be more desirable when interpretability is much more important than accuracy, or when interpretable models trained on a small number of carefully engineered features are as accurate as black-box models.

Another challenge for model-agnostic explanations is to be actionable.
Using a white box makes it easier to incorporate user feedback in systems like iBCM~\cite{kim2015ibcm}, or injecting logic into matrix factorization~\cite{logicmf:naacl15}. 
Feature labeling~\cite{druck2008learning} or annotator rationales~\cite{zaidan-eisner:2008:gen} are other forms of feedback that should be supported for explanations. 
A basic form of feature engineering (removing bad features) via explanations has been shown to be effective~\cite{lime}, but incorporating more powerful forms of feedback from the users is still a challenging research direction, in particular while remaining model-agnostic.

\section{Local Interpretable Model-agnostic Explanations (LIME)}

\cut{
\begin{figure}[!t]
\centering
\subfigure[Original Image]{
\includegraphics[width=0.22\textwidth]{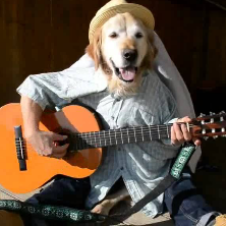}
\label{fig:original}}
\subfigure[Explaining \emph{Electric guitar}]{
\includegraphics[width=0.22\textwidth]{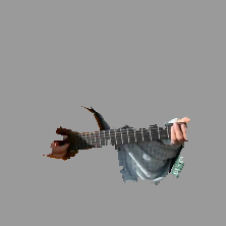}
\label{fig:electric}}
\subfigure[Explaining \emph{Acoustic guitar}]{
\includegraphics[width=0.22\textwidth]{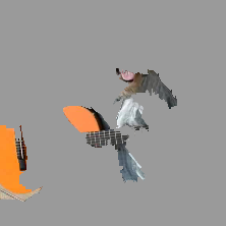}
\label{fig:acoustic}}
\subfigure[Explaining \emph{Labrador}\label{fig:labrador}]{
\includegraphics[width=0.22\textwidth]{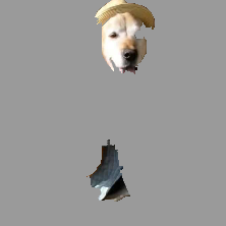}}
\caption{Explaining an image classification prediction made by Google's Inception network, highlighting positive pixels. The top 3 classes predicted are ``Electric Guitar'' ($p=0.32$), ``Acoustic guitar'' ($p=0.24$) and ``Labrador'' ($p=0.21$) }
\label{fig:inception}
\end{figure}
}

We now briefly review LIME~\cite{lime}, and discuss how it maintains model-agnosticism, while addressing some of the challenges that are described in the previous section.
We denote $x \in \mathbb{R}^d$ as the original representation of an instance being explained, and we use $x' \in \mathbb{R}^{d'}$ to denote a vector for its interpretable representation.
As exemplified before, $x$ may be a feature vector containing word embeddings, with $x'$ being the bag of words.

LIME's goal is to identify an \textbf{interpretable} model over the \emph{interpretable representation} that is \textbf{locally faithful} to the classifier.
Even though an interpretable model may not be able to approximate the black box model globally, approximating it in the vicinity of an individual instance may be feasible. 
Formally, the explanation model is $g: \mathbb{R}^{d'}\rightarrow\mathbb{R}, g \in G$, where $G$ is a class of potentially interpretable models, such as linear models, decision trees, or rule lists, i.e.
given a model $g \in G$, we can present it to the user as an explanation with visual or textual artifacts. 
As noted before, not every $g \in G$ is simple enough to be interpretable - thus we let $\Omega(g)$ be a measure of complexity (as opposed to interpretability) of $g$, which may be either a soft constraint (e.g. the depth of a tree, or the number of non-zeros in a linear model) or a hard constraint (e.g. $\infty$ if the depth or the number of non-zeros is above a certain threshold). 

\begin{figure}[tb]
\centering
\includegraphics[width=0.6\columnwidth]{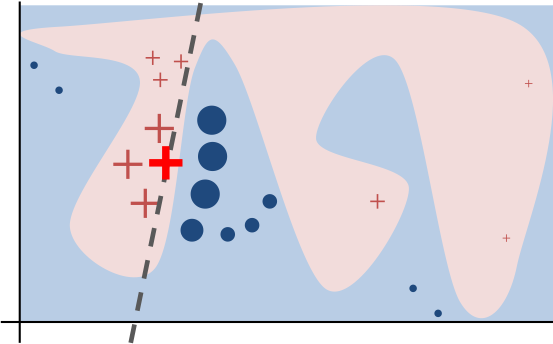}
\caption{Toy example to present intuition for LIME. The black-box model's
complex decision function $f$ (unknown to LIME) is represented by the blue/pink
background. 
The bright bold red cross is the instance being explained. LIME samples instances, gets predictions using $f$, and weighs them by the proximity to the instance being explained (represented here by size). The dashed line is the explanation that is locally (but not globally) faithful.}
\label{fig:lime}
\end{figure}

Let the model being explained be $f: \mathbb{R}^d\rightarrow \mathbb{R}$, e.g. in classification $f(x)$ is the probability that $x$ belongs to a certain class. 
We further use $\Sim_{x}(z)$ as a proximity measure between an instance $z$ to $x$, so as to define locality around $x$.
Finally, let $\mathcal{L}(f, g, \Sim_x)$ be a measure of how unfaithful $g$ is in approximating $f$ in the locality defined by $\Sim_x$. 
In order to ensure both interpretability and local fidelity, we must minimize $\mathcal{L}(f, g, \Sim_x)$ while having $\Omega(g)$ be low enough to be interpretable by humans.
The explanation $\EModel(x)$ produced by \textbf{LIME} is obtained by solving:
\begin{equation}
\EModel(x) = \argmin_{g \in G}\;\;\mathcal{L}(f, g, \Sim_x) + \Omega(g)
\label{eq:lime}
\end{equation}
This formulation can be used with different explanation families $G$, fidelity functions $\mathcal{L}$, and complexity measures $\Omega$.
We estimate $\mathcal{L}$ by generating perturbed samples around $x$, making predictions with the black box model $f$ and weighting them according to $\Sim_x$.
The intuition for this is presented in Figure \ref{fig:lime}, where a globally complex model is explained using a locally-faithful linear explanation.


\subsection*{Discussion}
Some approaches are model agnostic by approximating the black box model by an interpretable one globally~\cite{craven,Baehrens:2010:EIC:1756006.1859912,explain:krr15}.
Global explanation, however, are often either not interpretable, or too simplistic to represent the original model.
LIME's focus on explaining individual predictions allows more accurate explanations while retaining \textbf{model flexibility}. 
For example, it is easy to explain why sentences such as ``This is not bad.'' have a positive sentiment, even if we are not able to explain the complete sentiment model.

For \textbf{explanation flexibility}, the practitioner has complete control over $G$ and $\Omega(g)$;
in \citet{lime}, for example, we use very sparse linear models. 
This representation is simple enough for non-expert Mechanical Turkers to perform model selection and feature engineering effectively for complex, uninterpretable models. 
Furthermore, since LIME estimates the local fidelity through $\mathcal{L}$, we can directly control the interpretability of the explanations (e.g. using as many words as needed to maintain faithfulness) or whether to only display interpretable explanations when they are accurate to the black box model.
LIME also supports exploring multiple explanation families $G$ simultaneously, and picking the one with highest faithfulness. 

\textbf{Representation flexibility} is built into LIME, with the distinction between original $x$ and interpretable representation $x'$.
In \citet{lime}, we explain models trained on on word embeddings by using words as interpretable representation, and a neural network trained on raw pixels by using contiguous super-pixels as $x'$.

We demonstrate the small \textbf{switching costs} of LIME by explaining a wide variety of models (random forests, SVMs, neural networks, linear models, and nearest neighbors) using the same type of explanations. 
We also demonstrate LIME's utility for model comparison by enabling non-expert Mechanical Turk users to select which of two competing models would generalize better using the explanations. 

\begin{figure}[!t]
\centering
\subfigure[Logistic Regression trained on unigrams]{
\includegraphics[height=80px]{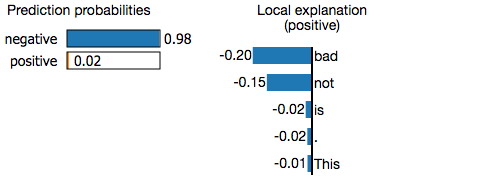}
\label{fig:logreg}}
\subfigure[LSTM trained on sentence embeddings.]{
\includegraphics[height=80px]{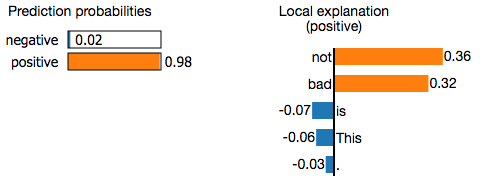}
\label{fig:lstm}}
\caption{Explaining sentiment predictions for the sentence ``This is not bad.'', using different models and representations}
\label{fig:comparing}
\end{figure}

As a final illustration, we explain the predictions two sentiment analysis classifiers on the sentence ``This is not bad.'', using the class of linear models as $G$. 
The classifiers vary wildly in complexity and underlying representation - one is a logistic regression trained on unigrams, while the other an LSTM neural network trained on sentence embeddings~\cite{wieting2016iclr}.
Explanations, given in terms of words (and their associated weights in a bar chart) in Figure \ref{fig:comparing}, demonstrate that completely different classifiers can be described in a unified, interpretable manner.
In Figure \ref{fig:lstm}, the explanation assigns positive weight to both ``not'' and ``bad'', as only the conjunction is responsible for the LSTM's positive prediction (even though interactions are not modeled explicitly).

\section{Conclusion}

Although interpretable models provide crucial insight into why predictions are made, they impose restrictions on the model, representation (features), and the expertise of the users.
We argued that model-agnostic explanation systems provide a generic framework for interpretability that allows for flexibility in the choice of models, representations, and the user expertise. 
We outlined a number of challenges that need to be addressed for model-agnostic approaches; some of which are addressed by the recently introduced LIME~\cite{lime}, while others are left as future work.
We thus conclude that model-agnostic interpretability is a key component in making machine learning more trustworthy - and ultimately, more useful.



\clearpage

\section*{Acknowledgements}


This work was supported in part by ONR awards \#W911NF-13-1-0246 and \#N00014-13-1-0023, and in part by TerraSwarm, one of six centers of STARnet, a Semiconductor Research Corporation program sponsored by MARCO and DARPA.


\bibliography{references}
\bibliographystyle{icml2016}

\end{document}